\documentclass{article}
\usepackage{spconf,amsmath,graphicx}
\usepackage{cite}

\title{Dynamic Object Removal for Effective SLAM}
\name{Phani Krishna Uppala\sthanks{School of Computer Science}, \;\;Abhishek Bamotra\sthanks{College of Engineering}, \;\;Raj Kolamuri\footnotemark[2]}
\address{  }

\begin{document}
%
\maketitle
\begin{abstract}
This research paper focuses on the problem of dynamic objects and their impact on effective motion planning and localization. The paper proposes a two-step process to address this challenge, which involves finding the dynamic objects in the scene using a Flow-based method and then using a deep Video inpainting algorithm to remove them. The study aims to test the validity of this approach by comparing it with baseline results using two state-of-the-art SLAM algorithms, ORB-SLAM2 and LSD, and understanding the impact of dynamic objects and the corresponding trade-offs. The proposed approach does not require any significant modifications to the baseline SLAM algorithms, and therefore, the computational effort required remains unchanged. The paper presents a detailed analysis of the results obtained and concludes that the proposed method is effective in removing dynamic objects from the scene, leading to improved SLAM performance.
\end{abstract}
\section{Introduction}
\label{sec:intro}

Since dynamic objects pose a challenge to effective motion planning and localization, we study the extent of this challenge and explore potential solutions. This solution works by removing the dynamic objects from the scene. Towards this goal, we follow a two-step process, (i) First, we aim to find the dynamic objects in the scene using a Flow-based method. (ii) Following that, we use deep Video inpainting \cite{kim2019deep, Kundu_2018_CVPR, Uppala-2021-129108} algorithm for removing the dynamic actors \& objects from the scene. To test this method's validity, we use two state of the art slam algorithms, ORB-SLAM2 and LSD. Comparing our approach with baseline results, we seek to understand the impact of dynamic objects and the corresponding trade-offs.

Our approach transforms the data without any significant modifications to the baseline SLAM algorithms. Hence computational effort required is unchanged. Deep learning based approaches have established new state of the art in various tasks\cite{dyn_obj, liu2019coherent, 8543868} have We use the recent advancements in the deep learning based network architectures

\section{Literature review} \label{LiteratureReview}
\subsection{Inpainting} \label{lr_inpainting}
Inpainting has wide-ranging applications, most well known among them being photo editing, re-targeting, object removal, etc. In recent years a wide variety of inpainting techniques have been explored. Generative approaches like deep image prior\cite{Ulyanov_2020} use convolutional neural networks to reconstruct the image by masking out the inpainted part from the image. The feature reuse of the conv nets\cite{8658966} forces the inpainted part of the image to be close to the rest of the original image.
Inpainting approaches have also incorporated attention to improving the network focus on the area to be inpainted\cite{liu2019coherent, xie2019image, Mopuri_2018_ECCV}.
Recently Nvidia is offering inpainting as a web service, in which non-technical users can directly upload the image and draw lines to mark the part to be inpainted. This clearly shows production-level performance that can be achieved using the state of the art inpainting techniques. We want to exploit these developments in Inpainting to study the effects of dynamic objects on localization and mapping problems.

Category-specific inpainting methods have also been developed towards inpainting\cite{lahiri2019faster}, this can be especially useful for the removal of the vehicles/cars from the image as they form a majority of the dynamic objects.

As the datasets for SLAM may not contain the necessary supervision required for training an inpainting algorithm, we also researched the unsupervised approaches for inpainting\cite{kim2019deep}. These unsupervised approaches alleviate the requirement of training data and are usually trained in an adversarial manner.

Since video data has temporal information, using video data for inpainting compared to image data provides a much richer set of cues. Especially temporal cues that are completely missing in the image level inpainting\cite{kim2019deep}. These video approaches use flow-based information to recover the missing portions.
\subsection{ORB-SLAM}
ORB-SLAM \cite{ORBSlam}
is considered a state-of-the-art algorithm in the field of visual SLAM using monocular cameras. It can handle all the major tasks of SLAM: tracking, mapping, relocalization, and loop closing. It runs in real-time, in small as well as large environments, and both in indoor and outdoor environments. The following are some of the distinctive features of this approach:

\begin{itemize}
  \item It uses ORB, a 256-bit feature descriptor in its analysis. ORB finds the right balance for real-time processing speeds and rotational invariance required in the subsequent parts of the analysis. 
  \item It can process the data in real-time while accomplishing the tasks of tracking, local mapping, and loop closing in parallel. 
  \item A covisibility graph is used to contain information of similarity of mapping points between two frames observed. 
  \item A bag of words place recognition module is first created out of offline data and is used to perform loop detection and relocalization tasks. 
  \item It also has the distinctive feature of automatic map initialization without human intervention by selecting one out of either planar homography or non-planar fundamental matrix calculation heuristically. 
  \item In the thread of tracking, first ORB features are extracted, then the pose of the camera is calculated either from the previous frame using a motion model or using global relocalization techniques, and a local map is then constructed and revised if certain conditions are met. 
  \item In the thread of local mapping, new keyframes are created and inserted, and map points are created by triangulation techniques and then added to the map after bundle adjustment optimization. 
  \item In the thread of loop closing, the keyframes are put through several tests to check if they qualify for loop closing; similarity transformation is carried on the selected candidates to calculate the error being accumulated, and then loop fusion is carried out along with essential graph optimization technique to obtain better results. 
  \item The proposed algorithm is evaluated on 27 sequences from the most popular datasets, and its effectiveness in regular SLAM tasks is proved. 
\end{itemize}

\subsection{ORBSLAM2}

ORBSLAM2 ~\cite{ORBSlam2} builds upon ORB-SLAM by introducing new functionalities and works with stereo and RGBD cameras. Several changes have been suggested to the original monocular based ORB slam to incorporate and exploit stereo/depth information. 
\begin{itemize}
    \item Monocular, close and far stereo keypoints: Initially, stereo and RGBD key points are considered and then classified into close and far keypoints based upon thresholding depth information. 
    \item Close keypoints can be safely triangulated from one frame, and it takes multiple frames to triangulate far points. Monocular keypoints, however, are defined at the points where depth information cannot be obtained both from stereo or RGBD. 
    \item This paper uses depth information from stereo or RGBD cameras to create a keyframe right from the first frame itself and use all stereo points to create an initial map.
    \item The obtained stereo keypoints are added in bundle adjustment for optimizing the camera pose in the tracking thread. 
    \item During loop closing, unlike monocular slam, the scale is now observable in ORBSLAM 2, and this information can help in geometric validation and pose-graph optimization without dealing with scale-drift.
    \item During Keyframe Insertion, the information of distinction between close and far stereo points is used to introduce another new condition for keyframe insertion, which is basically keeping a threshold for insertion based on the number of close and far keypoints. 
    \item  A new kind of localization is introduced, which tried to find matches between ORB in the current frame and 3D points created in the previous frame from the stereo/depth information making the localization of camera robust to unmapped regions. 
    \item The proposed algorithm delivers good results on 29 popular datasets with reduced translation RMSE. The authors also claim that this algorithm is the best SLAM solution for the KITTI visual odometry benchmark. 
\end{itemize}

\subsection{LSD-SLAM}

\begin{itemize}
    \item LSD-SLAM \cite{LSDSlam} is another state-of-the-art algorithm in the stream of visual SLAM that also uses monocular cameras. 
    \item It uses direct image intensities instead of feature descriptors to perform tracking and mapping operations. 
    \item The camera's pose is estimated using direct image alignment, and semi-dense depth maps are used to estimate 3D geometry. 
    \item A pose-graph of keyframes approach is used to build scale-drift corrected, large-scale maps and loop-closures. 
\end{itemize}

\section{THEORY/METHOD}

We propose an integrated framework for the stated objective of dynamic object removal for effective SLAM. First we generate masks in each frame, of the dynamic moving objects using Yang et al.\cite{yang2019unsupervised} We then use Inpainting algorithm to remove these dynamic objects from the scene. We test the implementation of this removal by comparing the results obtained on standard SLAM systems, namely ORB SLAM2 and LSD SLAM with and without dynamic objects.

\subsection{Dynamic Object Detection} \label{Dynamic Object Detection}

As an alternative to ORB feature-based localization of dynamic objects, we used Unsupervised Moving Object Detection via Contextual Information Separation ~\cite{yang2019unsupervised}. This is an optical flow-based method that works on a non-static camera motion. When we ran the algorithm on the benchmarking dataset for video object segmentation, Davis dataset\cite{davis_dataset}. We were able to replicate qualitative and quantitative results, as reported in \ref{Midterm_results}. This is a recent CVPR '19 publication; we made changes to the official code and approach to get results on the Kitti dataset. These modifications are primarily required for two main reasons. First, the lack of data augmentations proposed by this approach on the KITTI dataset. We resolved this by adding our own custom implementation. Second, official implementation assuming ground truth availability for the dataset. This is not the case with the KITTI dataset; hence, we reformulated the testing graph without the ground truth dependency by modifying the existing code. The results for the same are provided in section \ref{Midterm_results}. 

Based on the dynamic object detection algorithm results, we are taking the generated masks for the Kitti dataset and performed inpainting using Deep Video Inpainting ~\cite{kim2019deep}. The results for the inpainting are shown in section \ref{Midterm_results} as well. 

\subsection{Dynamic Object Removal: InPainting} \label{InPainting}

As a way to remove the dynamic objects from the input sensor data, we did a thorough literature survey on various inpainting algorithms and presented them in section \ref{lr_inpainting}. Of these algorithms, we tested the two most promising approaches. 

\begin{itemize}
    \item We tried two approaches: 1) Image based inpainting approach, 'Deep image prior' 2) video based inpainting, 'Deep video inpainting'.
    \item Upon evaluation of these two algorithms, the time taken by deep image prior using a single 1080Ti GPU is quite higher than that of deep video inpainting. Also, as the input modality of KITTI and TUM RGBD is video, we use deep video prior from here on.
    \item We independently tested our inpainting results on a standard inpainting benchmarking dataset(Davis) by inpainting over the segmentation masks.
    \item For dynamic object removal, we faced a technical challenge of high sparsity in features (ORB) to achieve localization.
    \item We explored different approaches to resolve this and localize dynamic objects. Moreover, we used the Contextual Information Separation ~\cite{yang2019unsupervised}. This approach is elaborated more below.
    
\end{itemize}

\subsection{ORB SLAM 2 and LSD SLAM} \label{ORB SLAM 2 and LSD SLAM}

We did an in-depth literature review of ORB SLAM2 and LSD SLAM. We give a brief overview of these papers in the Literature Review section \ref{LiteratureReview}. 
We performed baseline implementations of both ORB SLAM2 and LSD SLAM. We were able to obtain visualizations of the same. We implemented ORB SLAM2 on the TUM RGBD and Kitti dataset and implemented LSD SLAM on the TUM RGBD and TUM room sequence datasets. 


\section{Datasets}
We have performed  our experiments on KITTI \cite{Geiger2012CVPR}and TUM-RGBD datasets \cite{sturm12iros}. We used the same datasets and the same algorithmic setting for comparison across algorithms. Thus performing a controlled \& fair experiments on all the datasets. Description of various datasets in our pipeline are detailed below.

\subsection{DAVIS 2016}

Densely Annotated Video Segmentation (DAVIS)\cite{Perazzi2016} is a video object segmentation dataset; this is released as part of the DAVIS challenge. We used the most common variant of DAVIS 2016 for both of our inpainting and dynamic object segmentation tasks. This public bench marking dataset contains dynamic scenes with complex camera movements, making it realistic. This dataset also provides rich annotation in terms of pixel-wise segmentation mask. Our trained models for inpainting\cite{kim2019deep} used the segmentation as mentioned above the mask. Following\cite{kim2019deep} we used the annotations available on this dataset to independently test the inpainting model before using the results for SLAM. These results are reported in the results section. We followed the same approach of independent evaluation for dynamic object segmentation and presented the results.

\subsection{KITTI dataset}

The odometry benchmark from the KITTI dataset consists of 22 stereo sequences, saved in loss less png format. It provides 11 sequences (00-10) with ground truth trajectories, and we will use those for our project. The data is collected from a car driven around a residential area with accurate ground truth from GPS and a Velodyne laser scanner. This is a very challenging dataset for monocular vision due to fast rotations, areas with much foliage, which make more difficult data association, and relatively high car speed, being the sequences recorded at 10 fps.\cite{Geiger2012CVPR}
\subsection{TUM-RGBD dataset}

We will also test our results with the same parameters on the TUM-RGBD data. The dataset contains the color and depth images of a Microsoft Kinect sensor along the ground-truth trajectory of the sensor. The data was recorded sensor resolution (640x480) at full frame rate (30 Hz). The ground-truth trajectory is provided with the dataset and was obtained from a high-accuracy motion-capture system with eight high-speed tracking cameras with high frequency (100 Hz). The data also comes with accelerometer data recorded by the Kinect sensor.\cite{sturm12iros}

\section{Code: Technical Implementation}

Since the odometry split of the KITTI dataset is without segmentation, existing dynamic objection detection approaches do not benchmark on this dataset\cite{dyn_obj}. The approach we used\cite{dyn_obj} did the same. The workaround this we implemented our own dataloader in the format that is expected as of \cite{dyn_obj}. This implemented code is combined with the existing dynamic object detection code. Along with the above challenge, the KITTI odometry split does not have the ground truth corresponding to dynamic objects. To address this, we re-implemented the testing graph by taking parts of the existing code.
The inpainting algorithm, deep video inpainting, uses a deep network for effective inpainting. Towards this code setup, we used GPU packages, including Cuda toolkit, PyTorch, and gcc5. We used 4 Nvidia 1080Ti for our inference.
Similarly, our dynamic object segmentation approach is a neural network based model as well. Towards setting up this codebase, we installed GPU compatible TensorFlow, Keras, Cuda toolkit, and cudnn. We used 2 Nvidia 1080Ti for the inference.


\section{Results}

\subsection{Dynamic object detection}

\begin{figure}[h]
    \centering
    \includegraphics[width=0.35\textwidth]{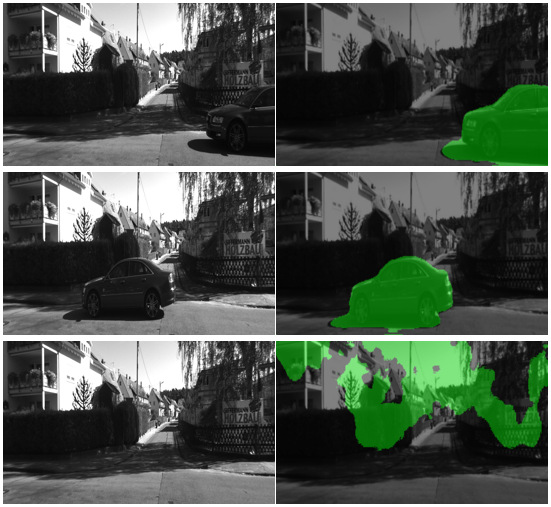}
    \caption{Results for dynamic object detection on the Kitti dataset. Images on the left show the original images. Images on the right show the corresponding original images with detected mask overlay.}
    \label{fig:mask}
\end{figure}

The results for dynamic object detection algorithm are shown in Fig. \ref{fig:mask}. From the results, we can see that the dynamic object detection algorithm, based on an optical flow of features can get good results and can be visually confirmed by the examples shown. However, there can be failure cases as shown in Fig. \ref{fig:mask} (last example). In that example, no dynamic object is detected; still, the algorithm produces the resultant mask. Such issues are observed when the car is taking a turn or suddenly changes its direction.

\subsection{Inpainting} 
\begin{figure}
    \centering
    \includegraphics[width=0.48\textwidth]{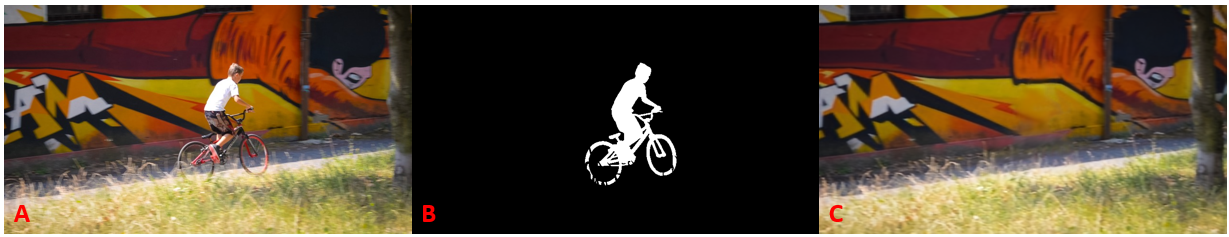}
    \caption{Inpainting result on the benchmarking video object segmentation data, Davis dataset. Fig. \ref{fig:inpainting_davis} A) Shows the original image from which the cyclist is removed. Fig. \ref{fig:inpainting_davis} B) Shows the mask for the object to remove. Fig. \ref{fig:inpainting_davis} C) Showcases the inpainting result after removing the object.}
    \label{fig:inpainting_davis}
\end{figure}

\begin{figure}
    \centering
    \includegraphics[width=0.35\textwidth]{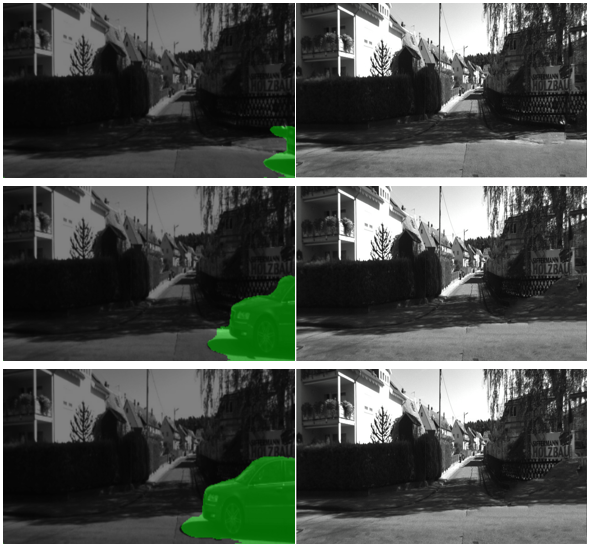}
    \caption{Results for inpainting on the KITTI dataset using the mask generated from the dynamic object detection algorithm. Images on the left show the original images with the green as mask for the dynamic object found in the frame. Images on the right show the corresponding inpainted version of the original images.}
    \label{fig:inpainting_kitti}
\end{figure}

The inpainting result for the benchmarking DAVIS dataset is shown in Fig. \ref{fig:inpainting_davis} and produces very good inpainting results using the algorithm mentioned before. We applied the same learned model on the KITTI dataset to remove the dynamic objects detected by the algorithm and were able to achieve good results for removing the object. There are some artifacts in some cases and result in distorted images, but most of the time, it works well. This is because the original model we are using to perform the inpainting is trained on RGB images, and the KITTI odometry dataset we are using is grayscale. Some examples are shown in Fig. \ref{fig:inpainting_kitti}. Following the previous inpainting algorithms, we measured the performance on Davis using the FID score, an inception based score to measure how realistic the generated images are.
\begin{table}[]
\centering
\caption{FID score for Davis dataset of 20 videos}
\begin{tabular}{|c|c|}
\hline
\textbf{Network} & \textbf{FID Score} \\ \hline
VINet (agg + T.C.) (in use) & \textbf{0.0046} \\ \hline
\end{tabular}
\end{table}
\begin{figure}
    \centering
    \includegraphics[width=0.45\textwidth]{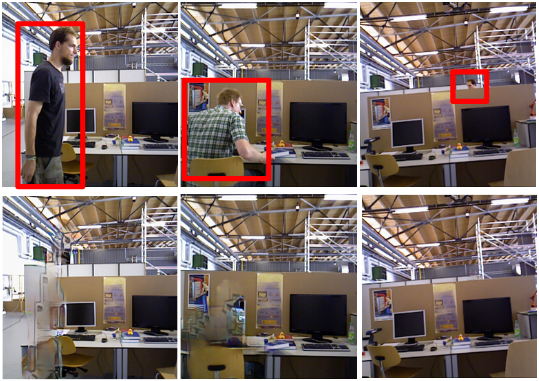}
    \caption{Results for inpainting on the TUM-RGBD dataset using the mask generated from the dynamic object detection algorithm. The images in the first row show the original image with a red bounding box for the dynamic object to be removed. The images in the bottom row show the inpainted results for the corresponding images.}
    \label{fig:my_label}
\end{figure}
\subsection{ORB-SLAM2}
\begin{figure}
    \centering
    \includegraphics[width=0.3\textwidth]{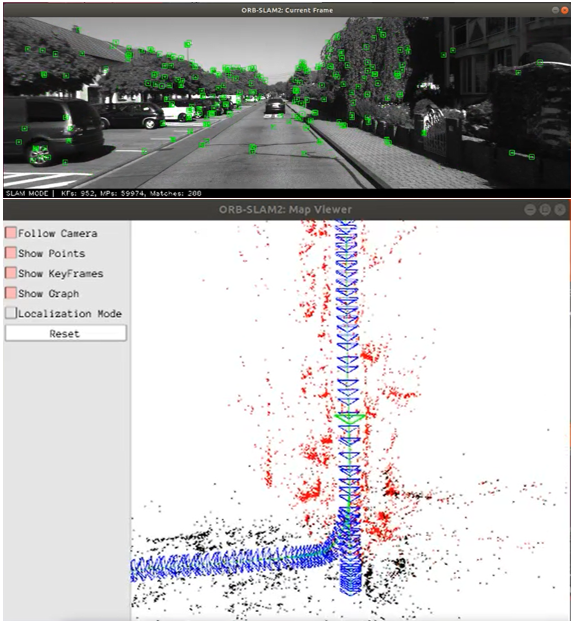}
    \caption{Result demonstrating loop closure for sequence 05 from Kitti dataset using ORB SLAM2.}
    \label{fig:orb_kitti}
\end{figure}

\begin{figure}
    \centering
    \includegraphics[width=0.4\textwidth]{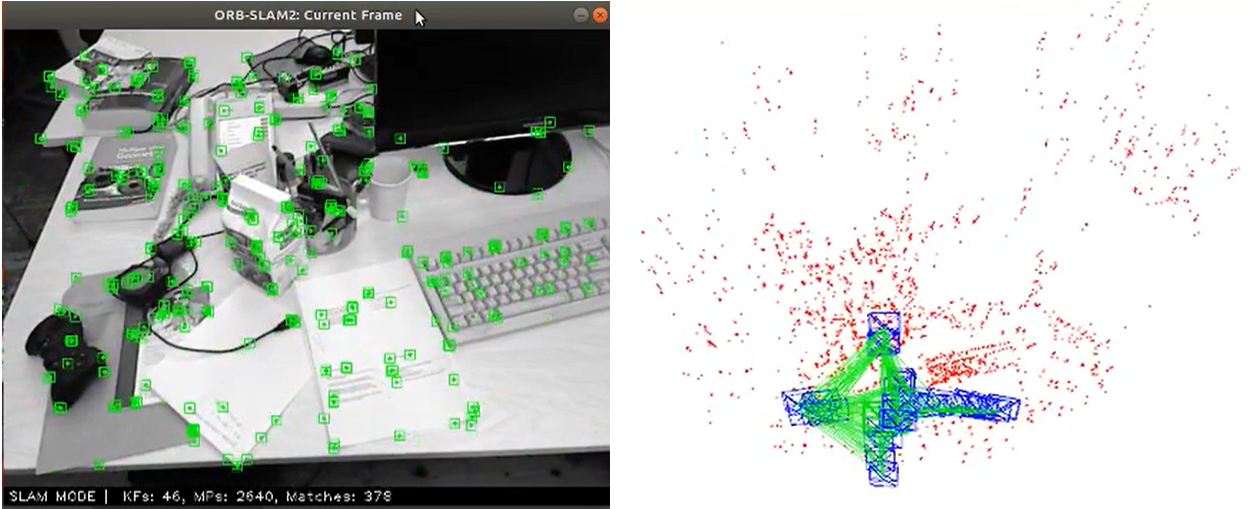}
    \caption{Resultant trajectory from the TUM-RGBD frb1xyz using ORB-SLAM2. Left image shows the camera image with orb features overlay and right image shows depth reconstruction and resultant trajectory.}
    \label{fig:orb_tum}
\end{figure}

The results for the ORB-SLAM2 on the KITTI data grayscale odometry data sequence 05 with loop closure is shown in Fig. \ref{fig:orb_kitti}. The result for TUM-RGBD frb1xyz using ORB-SLAM2 is shown in Fig. \ref{fig:orb_tum}. Reported results are estimated using the code and resources provided through the official implementation. These results for both the datasets look promising qualitatively as well.

\begin{figure}
    \centering
    \includegraphics[width=0.25\textwidth]{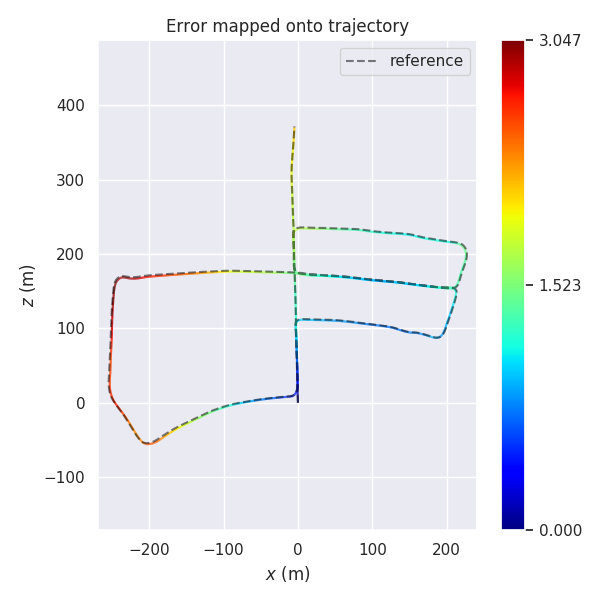}
    \caption{Absolute pose error color map KITTI 05 sequence without dynamic object for ORB-SLAM2}
    \label{fig:ape_plot}
\end{figure}

\begin{figure}
    \centering
    \includegraphics[width=0.25\textwidth]{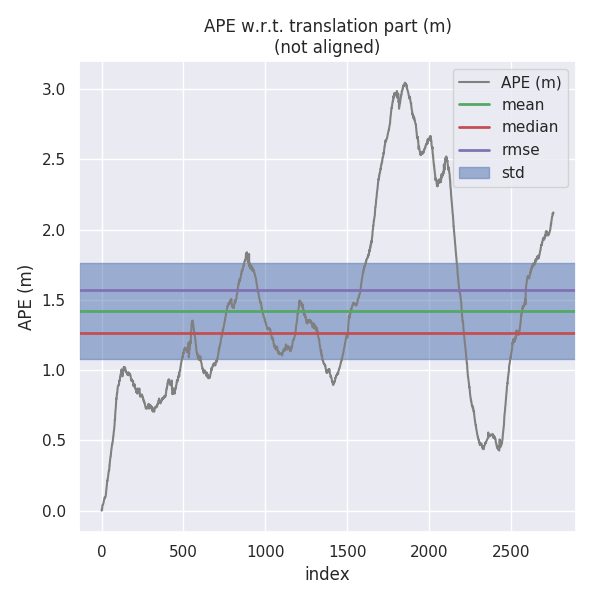}
    \caption{Absolute pose error with respect to the translation for KITTI 05 sequence without dynamic object for ORB-SLAM2}
    \label{fig:ape_color_path}
\end{figure}

\begin{figure}
    \centering
    \includegraphics[width=0.25\textwidth]{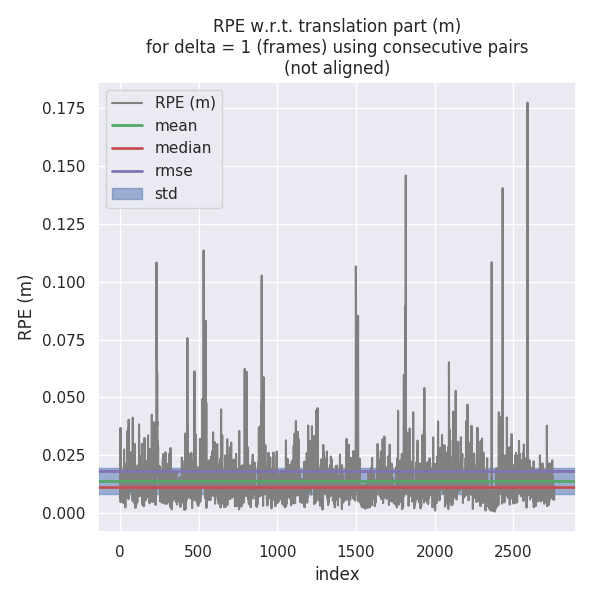}
    \caption{Relative pose error with respect to the translation for KITTI 05 sequence without dynamic object for ORB-SLAM2}
    \label{fig:rpe_plot}
\end{figure}
\begin{table}[]
\caption{RMSE Absolute Pose Error and Relative Pose Error for baseline (with dynamic objects) vs without dynamic objects for KITTI Sequences 00-07 with ORB-SLAM2}
\label{tab:KITTI_result}
\resizebox{0.48\textwidth}{!}{%
\begin{tabular}{|l|l|l|l|l|}
\hline
\textbf{Sequence} & \textbf{\begin{tabular}[c]{@{}l@{}}APE\\ (Baseline)\end{tabular}} & \textbf{\begin{tabular}[c]{@{}l@{}}RPE\\ (Baseline)\end{tabular}} & \textbf{APE} & \textbf{RPE} \\ \hline
00                & 2.669527                                                          & 0.301004                                                          & 2.706672     & 0.302460     \\ \hline
01                & 5.843165                                                          & 0.763168                                                          & 6.505854     & 0.795611     \\ \hline
02                & 1.856006                                                          & 0.024530                                                          & 2.445231     & 0.037193     \\ \hline
03                & 3.059412                                                          & 0.529403                                                          & 2.987005     & 0.528847     \\ \hline
04                & 2.370536                                                          & 0.397730                                                          & 2.360059     & 0.414609     \\ \hline
05                & 1.564191                                                          & 0.016376                                                          & 1.574255     & 0.018023     \\ \hline
06                & 1.981206                                                          & 0.239740                                                          & 2.000956     & 0.250131     \\ \hline
07                & 1.726958                                                          & 0.030662                                                          & 1.798538     & 0.037644     \\ \hline
\end{tabular}%
}
\end{table}

\begin{table}[]
\caption{RMSE Absolute Pose Error (APE) and Relative Pose Error (RPE) for baseline (with dynamic objects) vs without dynamic objects for TUM-RGBD dataset with ORB-SLAM2}
\label{tab:tum_result}
\resizebox{0.48\textwidth}{!}{%
\begin{tabular}{|l|l|l|l|l|}
\hline
\textbf{Sequence}   & \textbf{\begin{tabular}[c]{@{}l@{}}APE\\ (Baseline)\end{tabular}} & \textbf{\begin{tabular}[c]{@{}l@{}}RPE\\ (Baseline)\end{tabular}} & \textbf{APE} & \textbf{RPE}      \\ \hline
Sitting\_rpy        & 3.258594                                                          & 0.012794                                                          & 3.241519     & 0.010996          \\ \hline
Sitting\_Static     & 3.477002                                                          & 0.005230                                                          & 3.477901     & 0.005188          \\ \hline
Sitting\_halfsphere & 2.693894                                                          & 0.007857                                                          & 2.910348     & 0.008991          \\ \hline
Walking\_halfsphere & 3.126076                                                          & 0.025338                                                          & 2.997350     & 0.022527          \\ \hline
Walking\_xyz        & 3.685020                                                          & 0.053205                                                          & 3.490609     & 0.050013          \\ \hline
Walking\_rpy        & 3.536281                                                          & 0.051050                                                          & 3.486296     & 0.041311 \\ \hline
\end{tabular}%
}
\end{table}

\subsection{LSD-SLAM} After struggling with so many issues because of ROS implementation, we were able to get results for LSD-SLAM on LSD-SLAM's TUM room sequence (Fig. \ref{fig:lsd_room}) and also on TUM-RGBD frb1xyz (Fig. \ref{fig:lsd_tum}) are shown below. However, we faced numerous issues with integrating our pipeline into LSD SLAM as there were multiple compatibility issues esp. since the SLAM system needs older versions of ROS and Linux. This made it very challenging to work and integrate. We hence made a trade-off to pursue to complete major objectives of the project instead of spending weeks on sorting out the issues. 
\begin{figure}[h]
    \centering
    \includegraphics[width=0.48\textwidth]{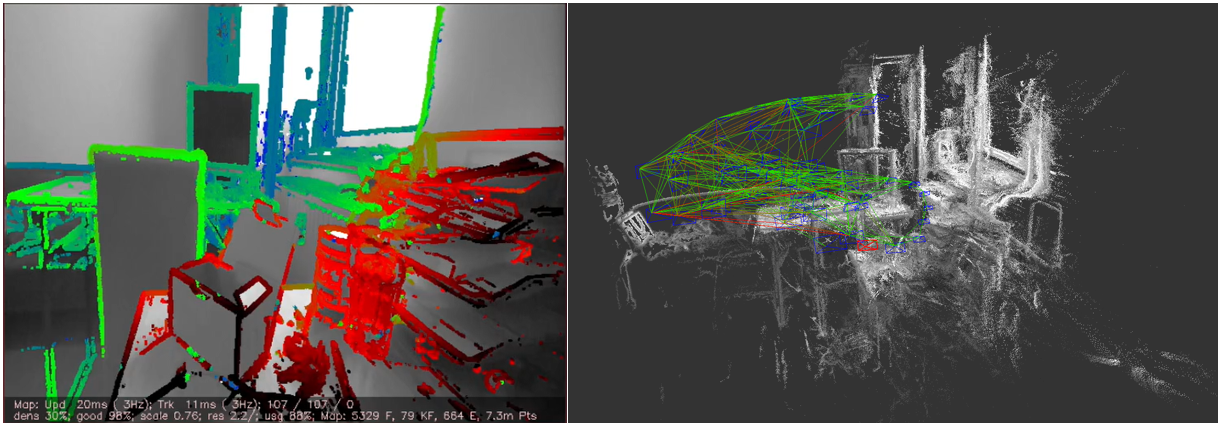}
    \caption{Result for LSD-SLAM on room sequence from TUM data is shown. Image on the right shows the pont cloud generated and the trajectory in green.}
    \label{fig:lsd_room}
\end{figure}

\begin{figure}[h]
    \centering
    \includegraphics[width=0.48\textwidth]{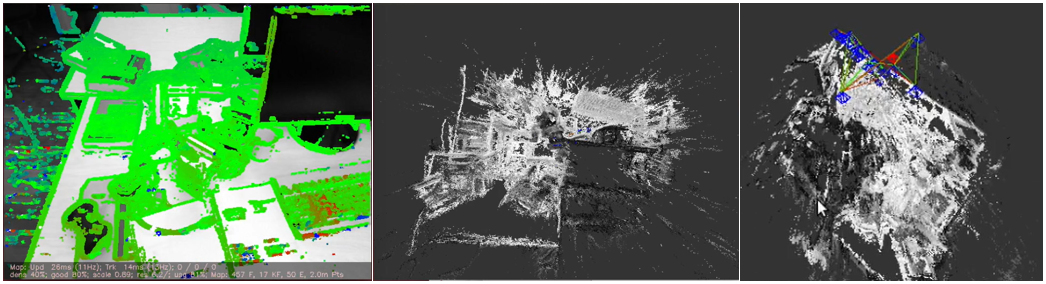}
    \caption{Result for LSD-SLAM on room sequence from TUM data is shown. Image on the right shows the point cloud generated and the trajectory in green.}
    \label{fig:lsd_tum}
\end{figure}

\subsection{Error Metrics}

Table \ref{tab:KITTI_result}  summarizes the error metrics, namely Absolute Pose Error(APE) and Relative Pose Error(RPE), with and without dynamic object removal on the KITTI dataset. We have shown the results for sequences 00-07. As can we be seen from the table, removing dynamic objects did not change the error metrics significantly. We believe this is happening because the relative no. of frames with dynamic objects in them is way smaller when compared to the TUM-RGBD dataset. For example, the ratio is $\approx$ 80:2800 frames which is way less for a good result in our pipeline. These errors can be visualized in Fig. \ref{fig:ape_plot}, \ref{fig:ape_color_path} and \ref{fig:rpe_plot}.

Table \ref{tab:tum_result} summarizes the error metrics, namely Absolute Pose Error(APE) and Relative Pose Error(RPE), with and without dynamic object removal. The table corresponds to the results on ORB SLAM2 system on TUM-RGBD dynamic sequences. It can observed from the table that both APE and RPE error metrics showed improvement generally. However, in cases like Sitting\_halfsphere dataset, we can see the error to be increased. When looked at the data, we found that it corresponds to a person sitting almost idle(with minute movements) with the camera having rigorous movement. We believe this affected the dynamic object detection algorithm with a noisy mask being generated resulting in higher error.

\subsection{Results against Expectations}
Initially, we planned to use ORB features for dynamic object detection and removal. However, we found it difficult to deal with these features due to their sparsity. Hence, we explored optical flow methods for dynamic object detection and removal(inpainting). We saw that removing dynamic objects from the scene showed minimal improvements in the error metrics like RMSE and RPE errors. Also, we observed that the proposed method is more effective for datasets which has more dynamic objects. This is intuitive because our method only deals with the removal of dynamic objects and doesn't deal with other facets in the scene. 

\section{Discussion}
\subsection{Sync with timeline}
As promised in our proposal, we were able to achieve majority of the stated objectives. Here are the specific objectives stated and their details of implementation. 
\subsubsection{ORB SLAM2 Implementation}
We completed doing baseline implementation of ORB SLAM2 on TUM RGBD \& KITTI datasets according to the schedule. 
\subsubsection{LSD SLAM Implementation}
We were able to complete the baseline implementation for LSD SLAM on TUM RGBD and TUM room sequence; according to the schedule.

\subsubsection{Dynamic Object Detection} Although we faced challenges for dynamic object detection through our initial proposed approach. We quickly explored other ways to localize dynamic objects. Through a thorough literature review in this sub field and few trails, we successfully identified a dynamic object detection that fits best for our task~\cite{yang2019unsupervised}. Using this method mentioned in Unsupervised Moving Object Detection via Contextual Information Separation ~\cite{yang2019unsupervised}, we were able to proceed experiments and attain both qualitative and quantitative results. 
\subsubsection{Inpainting}
We are able to successfully test two inpainting algorithms, Deep image prior \cite{Ulyanov_2020} and Deep video inpainting \cite{kim2019deep}. We were also able to do qualitative and quantitative evaluations of made both in terms of dynamic object detection.

\subsubsection{Dynamic Object Removal on ORB SLAM2}
We integrated the dynamic object removal into the ORB SLAM2 system on various datasets and compared how the errors metrics change when the dynamic objects are removed from the scene. 
\subsubsection{Dynamic Object Removal on LSD SLAM}
There were multiple challenges with implementing LSD SLAM on the recent versions of Ubuntu and ROS(LSD Slam being a ROS dependent system). Even though we could solve a lot of them, integrating our developed framework posed several complex challenges that rendered us not to pursue it further taking into account the investment of time vs. reaching project objectives.

\subsection{Technical Challenges} \label{tech_challenges}
\subsubsection{ORB-SLAM2-Implementation} For the ORB-SLAM2, we used the official implementation by the authors available on Github. We downloaded the source code for ORB-SLAM2, installed pre-required packages like Eigen3, OpenCV,  Pangolin (used for visualization), DBoW2, and g2o (third party), and C++11 or C++0x Compiler setup. We faced some issues with aligned and usleep while building CMake. To fix the issues, we included \#include $<$unistd.h$>$ in the system.h folder under include folder in ORB-SLAM2 master directory. The building of packages was successful, and we were able to get our results, as shown in the later section.

\subsubsection{LSD-SLAM-Implementation}
Initially, the LSD SLAM implementation given by TUM is ROS dependent. However, this implementation is based on either ROS Indigo or ROS Fuerte and with previous versions of Ubuntu like 12.04, 14. However, all our systems are configured to Ubuntu 18.04 and ROS Melodic. We initially tried to install it using some guidelines from others on Github using rosmake system on ROS Melodic itself. We faced many issues with versions of dependencies, and then after all of that, when the building started without errors, it kept building for hours(nearly 5 hours) without completion. We kept it building it for hours but with no luck. We then tried another implementation of LSD SLAM without the use of ROS. We also found a docker based implementation, but after doing all the steps mentioned in that, during the make phase, we got numerous errors. A number of them were correlated to the g20 package. Even after installing g20, multiple errors persisted. We then had to install and set up ros indigo on a new ubuntu 14.04 setup with no way around it. We were finally able to run the LSD-slam implementation on the above setup for 2 of the standard sequences. However the issues lingered and multiplied when we tried to integrate the entire framework and in the interest of the time, we focussed on other facets of the project.

\subsubsection{Inpainting}
Inpainting approaches work by taking a binary mask for the region to be inpainted. Since dynamic objects need to be inpainted for our task, a binary mask for each frame representing the dynamic objects would be required. Using these frame by frame masks, inpainting can be performed. However, no such mask data was available for dynamic objects in both TUM RGBD and Kitti datasets. As a way to resolve this, we explored approaches to localize dynamic objects. Moreover, we were able to obtain them and then use them for inpainting successfully.

\subsubsection{Dynamic object detection} 

Initially, we planned to use optical flow and our ORB features to generate a mask for our inpainting algorithm. After implementing the baseline for ORB SLAM2, we realized that the ORB features are quite sparse on the dynamic objects. There was no method to triangulate or find a convex hull over these dynamic objects. Therefore, we had to explore some other approaches to deal with this problem. 

We started with a literature review of existing approaches that localize dynamic objects. Most of these approaches assume a static camera or minimal camera motion, which is not the case for the datasets we are dealing with. To overcome this, we used one of the recent CVPR'19 works targeted to video object segmentation and re-purposed it to solve our problem.

We couldn't do a specific validation of the dynamic object detection as the ground truth labels are not available. However, we did perform a human visual inspection of the same as a validation procedure.

\section{Conclusion}
We integrated the dynamic object removal from the scene into ORB SLAM2 on TUM RGBD Dataset and KITTI Dataset. We showed how the evaluation parameters RMSE and Relative Pose Error(RPE) get affected by the dynamic object removal. However, due to multiple technical challenges involved, we could fully integrate the dynamic object removal into the LSD SLAM system. Overall, we observed that removing dynamic objects helps. Also, sequences with higher no. of dynamic objects show more sensitivity to their removal than the ones with less number of objects. We also observed that the dynamic object removal showed minimal improvement on the error metrics even though it didn't deteriorate. 

One possible extension of our work can be to check specifically how separate threads of tracking, mapping and localization get  affected by dynamic object removal. We can also test our approach on other state of the art SLAM algorithm esp. those that use dense features. It would be interesting to see how several error metrics get affected when the dynamic objects are removed. 


\bibliographystyle{IEEEbib}
\bibliography{strings,Template}
\end{document}